# Real-time windrow detection from onboard tractor sensors for automated following

Lorenz Gunreben, Nico Heider, Sebastian Zürner, Martin Schieck and Bogdan Franczyk[1]

**Abstract:** Proprietary design in commercial windrow-detection systems restricts transparency and limits progress in open autonomous forage-harvesting research. We present a multi-modal dataset combining stereo vision and LiDAR from tractor-mounted sensors during real baling operations. The dataset includes synchronized sensor data with GNSS trajectories, partly released as ROS2 Humble bags on Zenodo, with additional data available on request. Using this dataset, we implement a real-time (>20 Hz) centroid-based windrow-following method on an NVIDIA Jetson AGX Orin. Across the critical 4–10 m guidance range, stereo and LiDAR depth measurements show strong agreement ($0.965 \pm 0.021$), indicating that low-cost stereo sensors can approach LiDAR performance. Our open-source ROS 2 pipeline provides a reproducible benchmark for GPS-free windrow detection and supports development of practical autonomous forage-harvesting systems. Dataset: https://zenodo.org/records/17486318

**Keywords:** windrow detection, autonomous harvesting, multi-modal sensing, Stereo-LiDAR fusion, real-time processing, edge computing, agricultural dataset, ROS2

## 1    Introduction

Windrow detection is a critical perception task for autonomous forage harvesting.[2] During bale pressing, operators must continuously maintain precise alignment with windrows (long rows of cut and dried crop material) to ensure efficient pickup and high-quality bale formation. This repetitive task leads to operator fatigue and limits productivity, particularly during extended harvest periods.

Current commercial solutions for windrow guidance exist, but the underlying algorithms and training data remain proprietary, limiting scientific reproducibility. In academic literature, most studies on agricultural row detection focus on single sensor modalities and rarely release datasets or code, creating barriers to comparative research and sensor fusion development.

---

[1] Leipzig University, Smart Farming Lab, Grimmaische Straße 12, 04109 Leipzig, heider@wifa.uni-leipzig.de, gunreben@wifa.uni-leipzig.de, zuerner@wifa.uni-leipzig.de, schieck@wifa.uni-leipzig.de, francyk@wifa.uni-leipzig.de

[2] Language editing and text shortening were supported by AI-based tools (Claude, ChatGPT). All scientific concepts, methodology, and interpretations were developed by the authors.

We address these gaps by presenting a real-world, multi-modal dataset from tractor-mounted sensors (ZED2i stereo camera (Stereolabs, San Francisco, CA, USA) + Ouster OS0-128 REV7 LiDAR (Ouster Inc., San Francisco, CA, USA)) recorded during windrow baling operations, as well as a comparative analysis evaluating both modalities under field deployment conditions using inter-sensor agreement.

Real-time edge deployment on NVIDIA Jetson AGX Orin demonstrates >20 Hz processing with open-source ROS2 nodes. To our knowledge, this is the first publicly available dataset combining stereo and LiDAR for windrow detection, with both data and code released to establish reproducible baselines for GPS-free autonomous navigation in forage harvesting.

## 2     Related work

Vision-based crop row detection has been extensively studied for agricultural navigation. [Ch14] achieved 0.07 m lateral accuracy using laser scanners with cross-correlation, while [KFB20] demonstrated stereo-based crop edge detection with elevation mapping (0.09 m RMSE). [Yu21] developed ridge-furrow tracking using v-disparity stereo vision (2.5 cm accuracy), and [GSO23] used Velodyne LiDAR for ploughing furrow detection (86% agreement with annotations).

Most related, [Th23] demonstrated LiDAR-based windrow detection for compost turners using RANSAC ridge extraction, achieving 0.15 m standard deviation. However, their approach requires global point cloud reconstruction during an initialization phase and was validated only in simulation without real-world deployment.

A common limitation across prior work is the lack of publicly available datasets which prevents both reproducibility and the possibility of adapting algorithms.

## 3     Materials and methods

### 3.1     Dataset acquisition

To address the lack of publicly available agricultural machinery datasets, we established a continuous data collection program using a Fendt 724 tractor equipped with a multi-modal sensor carrier. The platform records diverse field operations throughout the growing season, capturing real-world farming workflows with the forward-facing Ouster OS0-128 LiDAR, ZED 2i stereo camera, and Novatel OEM7 GNSS (Hexagon AB, Stockholm, Sweden) for trajectory reference. Additional sensors (five ArkCam Basic+ Mini mono cameras, rear-mounted Blickfeld QB2 (Blickfeld GmbH, Munich, Germany) LiDAR) provide surround perception for other tasks but are not evaluated in this windrow detection study.

This paper focuses on a representative subset from forage harvest operations at Leipzig University's Oberholz research farm in Saxony, Germany. Field trials occurred under bright, dry conditions ideal for both sensing modalities. The LiDAR extrinsic parameters were manually set by aligning the sensor to the tractor CAD model, then the stereo camera was registered to the LiDAR frame via ICP alignment of overlapping point clouds.[3] This yielded sufficient accuracy within the windrow detection range. At longer distances, stereo depth uncertainty causes apparent lateral offsets in stationary objects, but near-field inter-sensor agreement remained reliable for comparison (Section 4.2).

### 3.2  Dataset description

The subset used for evaluation comprises 123 seconds capturing a single windrow under consistent conditions (35.9 GB, publicly available on Zenodo[4] as ROS2 Humble bag files). The full dataset spans 29 minutes with multiple windrows and turning maneuvers (262.2 GB, available upon request), forming part of a broader multimodal agricultural dataset initiative by our research group.

LiDAR and stereo data were recorded at 18.3 Hz (Ouster OS0-128, 131k points/frame) and 23.5 Hz (ZED2i, ~230k raw points/frame). The ZED point cloud was preprocessed to remove outliers and visual artifacts from lens wear, retaining an average of 66k points/frame after filtering.

### 3.3  Windrow detection algorithm

We employ a weighted centroid approach for centerline extraction, processing each longitudinal slice independently. For each row at distance y, cells exceeding an adaptive height threshold (p-th percentile, default $p = 0.5$) contribute to the centerline with weights proportional to their height above threshold. The lateral position is computed as a height-weighted average of cell positions.

This formulation biases toward elevated regions while maintaining robustness to noise and irregular geometries. Unlike top-K% selection [Th23], which can introduce lateral bias from asymmetric peak distributions, the weighted centroid balances all above-threshold contributions. Implementation details and code are available on GitHub.[5]

---

[3] Calibration files, urdf and sensor setup: https://github.com/Gunreben/vario700_sensorrig, 31.10.2025
[4] Zenodo: Windrow Detection Dataset, https://zenodo.org/records/17486318, 31.10.2025
[5] Windrow-Centerline-ROS2-Node: https://github.com/Gunreben/Windrow-Centerline-ROS2-Node, 31.10.2025

## 4 Results

### 4.1 Computational performance

The windrow detection node was evaluated on an NVIDIA Jetson AGX Orin using the 123-second dataset in Hardware-in-the-Loop configuration (server-streamed rosbag to NVIDIA Jetson). The system achieved real-time performance with mean processing times below 50 ms per frame (>20 Hz) without frame dropouts, demonstrating suitability for edge deployment on agricultural hardware. The results are visualized in Figure 1.

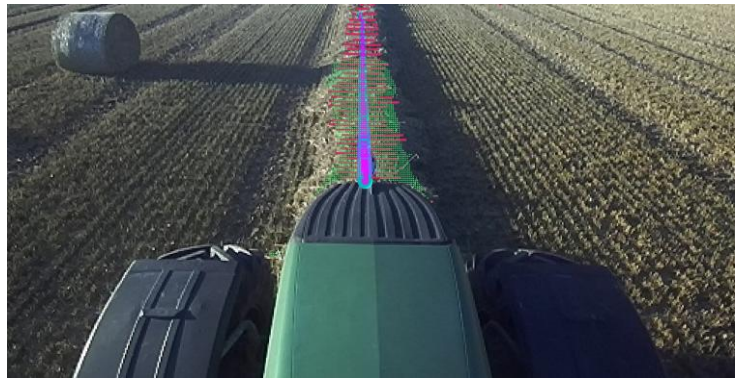

Fig. 1: Windrow centerline detection with multi-sensor fusion. Detected LiDAR centerline (magenta) and centerline points (blue) overlaid on front-camera view. Green: ZED2i stereo points; red: Ouster LiDAR points above 10cm height threshold

### 4.2 Inter-sensor agreement analysis

To evaluate detection consistency without ground truth, we computed inter-sensor agreement across 126 frames. The agreement score measures lateral offset between full centerline trajectories from both sensors:

$$Score = 1 - |\Delta x_{lateral}|$$

Where $\Delta x\_lateral$ is the mean lateral offset between sensor-specific centerlines in the vehicle frame.

The sensors demonstrated strong consistency (mean: 0.965 ± 0.021, median: 0.973). The three worst frames (out of the 126 analyzed frames) scored 0.904, 0.890, and 0.843, showing that cases above 10 cm lateral offset occur rarely. Both detected the windrow slightly left of center (ZED: −4.2 ± 13.4 cm; LiDAR: −6.3 ± 13.8 cm), likely from manual calibration bias or operator offset during acquisition.

Figure 2 shows best (0.998) and worst (0.843) agreement cases. High agreement occurs when centered and aligned with the windrow. While the worst frame (Fig. 2, right) suggests larger offsets at higher yaw angles, this was not confirmed. Some parallel centerlines showed high offsets, while others with stronger yaw did not. The tractor's roll angle or local surface unevenness may warrant further analysis.

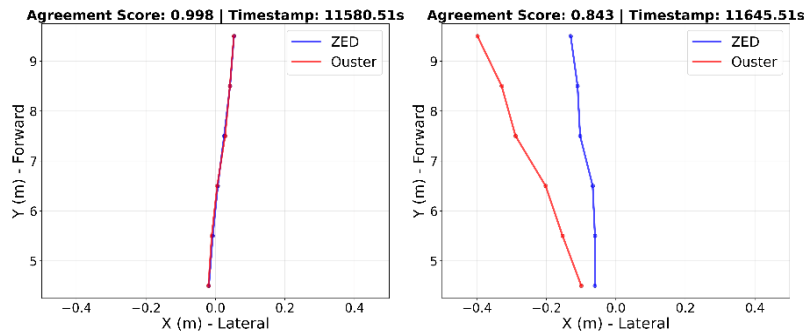

Fig. 2: Centerline detection comparison: best agreement (left, score=0.998) and worst agreement (right, score=0.843) between ZED (blue) and Ouster LiDAR (red) sensors.
Note the axis scales: lateral deviations span only ±0.5m over a 4-10m forward distance

## 5   Discussion

The high inter-sensor agreement (0.965 ± 0.021) demonstrates that cost-effective stereo vision (ZED2i, ~€450) provides windrow detection comparable to a high-resolution LiDAR (Ouster OS0-128, ~€8,000) for typical harvest operations. While lateral disagreement increases with distance due to stereo depth uncertainty, this degradation remains acceptable for near-field guidance (4-10m) at typical baler speeds (5-15 km/h), where both sensors agree closely.

Key deployment challenges include manual sensor calibration without factory integration and camera lens durability under field conditions (requiring 71% point filtering, due to lens wear). Future work will address multi-window scenarios with automated row switching and structured extrinsic calibration to reduce inter-sensor offsets. This dataset forms part of ongoing efforts to establish comprehensive multimodal benchmarks for agricultural perception.

## 6 Conclusion

This work establishes reproducible baselines for multi-modal windrow detection through three contributions: (1) the first open dataset combining stereo and LiDAR for windrow perception, publicly released with full ROS2 integration; (2) real-time edge deployment on Jetson AGX Orin achieving >20 Hz processing with inter-sensor agreement scores of 0.965 ± 0.021 and (3) practical insights on cost-performance tradeoffs and field deployment challenges. By releasing both data and code, we enable the research community to validate, extend, and improve upon these methods. The strong performance of cost-effective stereo cameras demonstrates feasibility for adoption in price-sensitive agricultural applications, while the dataset provides a foundation for advancing GPS-free autonomous navigation in forage harvesting.

## Acknowledgements

This work and the Rubin Feldschwarm® ÖkoSystem project are funded by the Federal Ministry of Research, Technology and Space (BMFTR) (grant no. 03RU2U051F, 03RU2U053C). We also thank NVIDIA for providing a Jetson AGX Orin through the NVIDIA Research Grant Program.